\documentclass[10pt,twocolumn,letterpaper]{article}

\usepackage{cvpr}
\usepackage{times}
\usepackage{epsfig}
\usepackage{graphicx}
\usepackage{amsmath}
\usepackage{amssymb}

\usepackage{algorithm}  

\usepackage{algpseudocode}
\usepackage{amsmath}
\newcommand{\tabincell}[2]{\begin{tabular}{@{}#1@{}}#2\end{tabular}}

\usepackage[pagebackref=true,breaklinks=true,letterpaper=true,colorlinks,bookmarks=false]{hyperref}

\cvprfinalcopy % *** Uncomment this line for the final submission

\def\cvprPaperID{2019} % *** Enter the CVPR Paper ID here

\ifcvprfinal\fi
\begin{document}

%%%%%%%%% TITLE
\title{Decorrelated Adversarial Learning for Age-Invariant Face Recognition}

\author{Hao Wang\quad Dihong Gong\quad Zhifeng Li\thanks{indicates corresponding authors.}\quad Wei Liu\footnotemark[1]\\
Tencent AI Lab  \\
\tt\small hawelwang@tencent.com,  gongdihong@gmail.com, michaelzfli@tencent.com, wl2223@columbia.edu
}
%\and

\maketitle

%%%%%%%%% ABSTRACT
\begin{abstract}
There has been an increasing research interest in age-invariant face recognition. However, matching faces with big age gaps remains a challenging problem, primarily due to the significant discrepancy of face appearances caused by aging. To reduce such a discrepancy, in this paper we propose a novel algorithm to remove age-related components from features mixed with both identity and age information. Specifically, we factorize a mixed face feature into two uncorrelated components: identity-dependent component and age-dependent component, where the identity-dependent component includes information that is useful for face recognition. To implement this idea, we propose the Decorrelated Adversarial Learning (DAL) algorithm, where a Canonical Mapping Module (CMM) is introduced to find the maximum correlation between the paired features generated by a backbone network, while the backbone network and the factorization module are trained to generate features reducing the correlation. Thus, the proposed model learns the decomposed features of age and identity whose correlation is significantly reduced. Simultaneously, the identity-dependent feature and the age-dependent feature are respectively supervised by ID and age preserving signals to ensure that they both contain the correct information. Extensive experiments are conducted on popular public-domain face aging datasets (FG-NET, MORPH Album 2, and CACD-VS) to demonstrate the effectiveness of the proposed approach.
\end{abstract}

%%%%%%%%% BODY TEXT
\section{Introduction}
Face recognition has been well studied for many years, with both traditional methods~\cite{liu2004null,li2005nonparametric,liu2006spatio,xiong2013face} and more recent deep learning based algorithms~\cite{deepid2,deepid2plus,deepid3,centerloss, cosface,amsm,sphereface,arcface} that have achieved excellent performance using deep learning networks such as~\cite{alexnet,vgg,ResNet101,resnext}. Many of these models are even more accurate than
humans in various scenarios. However, identifying faces
across a wide range of ages remains under-exploring.

\begin{figure}[t]
\begin{minipage}[b]{1\linewidth}
  \centering
  \centerline{\includegraphics[width=1.0\linewidth]{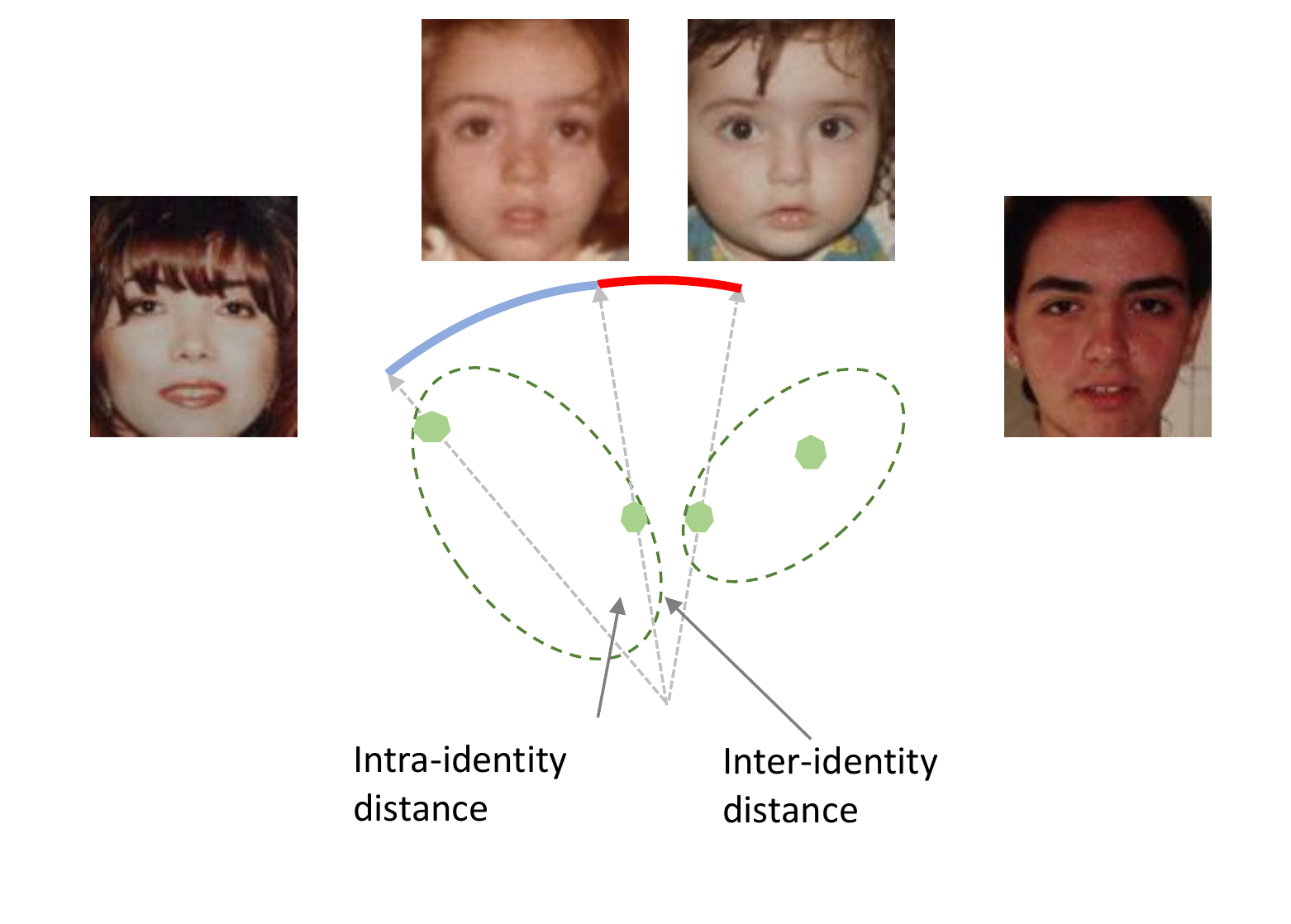}}
\end{minipage}
\caption{We show a typical example for AIFR, where the intra-identity distance is greater than the inter-identity distance due to the large age variations. As a result, many current face recognition systems fail to identify faces across big age gaps.}
\label{fig:intra_inter}
\end{figure}

Recently, modern advances \cite{normface,centerloss,sphereface,cosface,amsm,arcface} introduce the margin-based metrics and normalization mechanism to train the models in order to improve the face recognition performance. However, most of these methods usually lack the discriminating power for face  identification in the scenario of Age Invariant Face Recognition (AIFR). 
The crucial challenge for AIFR
is subject to the significant discrepancy resulting from the
aging process. 
Figure \ref{fig:intra_inter} shows an example that face images have great variations within the same identity across different ages, while those of different identities share  similar age-related information. As a result, those faces with big age gaps serve as hard examples that the current face recognition systems cannot identify correctly. In particular, the intra-identity distance is increasing larger if there are more faces of the child and the elderly.

In the meanwhile, increasing research attentions have been attracted to the age-invariant face recognition (AIFR). Recent research studies on AIFR mainly focus on the design of either generative models or discriminative models.
The generative methods \cite{g1,g2,g3} propose to synthesize face images of different ages to assist the face recognition. Very recently, several studies \cite{caae,cgan,TNVP} aim at improving the quality of generated aging faces by utilizing the powerful GAN-based models. However, accurately modeling the aging process is difficult and complicated. The unstable artifacts in the synthesized faces can significantly affect the performance of face recognition. In contrast, discriminative methods draw increasing interest in recent studies. For example , the \cite{hfa} separates the identity-related information and the age-related information through the hidden factor analysis (HFA). The \cite{LFCNN} is based on similar analysis and extends the HFA to the deep learning framework. More recently, the OE-CNN \cite{oecnn} presents the orthogonal feature decomposition to solve the AIFR. According to all these studies, feature decomposition plays a key role in invariant feature learning under the assumption that facial information can be perfectly modeled by the decomposed components. However, the decomposed components practically have latent relationship with each other and the identity-dependent component may still contain age information.

\begin{figure}[t]
\begin{minipage}[b]{0.9\linewidth}
  \centering
  \centerline{\includegraphics[width=1.0\linewidth]{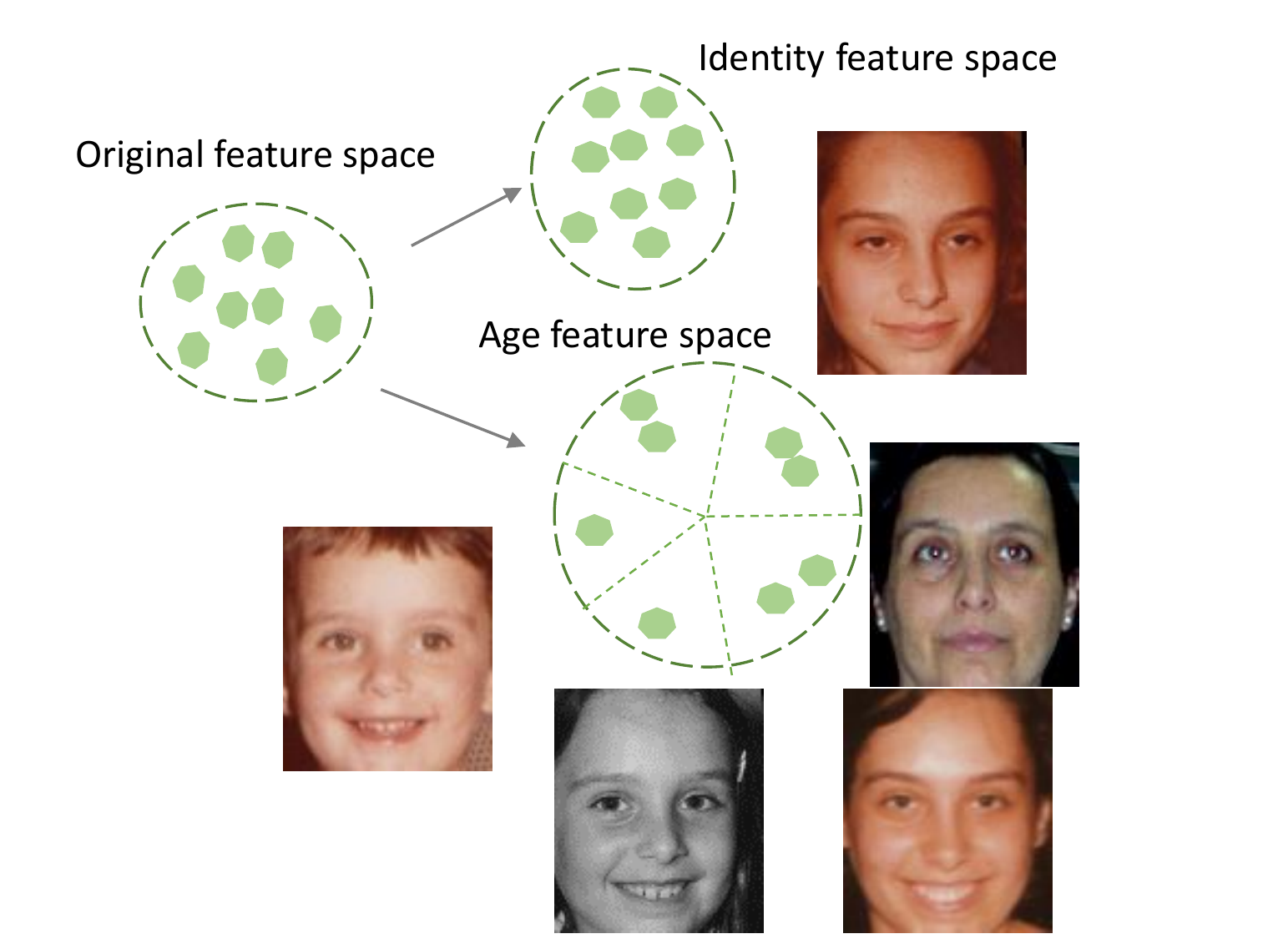}}
\end{minipage}

\caption{The face features are decomposed into the identity-dependent component and the age-dependent component. Only the identity features participate the testing of face recognition.}
\label{fig:space}
\end{figure}

In this paper, we introduce a deep feature factorization learning framework that factorizes the mixed face features into two uncorrelated components: identity-dependent component ($\mathbf{x_{id}}$) and age-dependent component ($\mathbf{x_{age}}$). Figure \ref{fig:space} illustrates our feature factorization schema. We implement such factorization through a residual mapping module inspired by \cite{dream}. This means that, the age-dependent embeddings are encoded through a residual mapping function $\mathbf{x_{age}}= \mathcal{R}(\mathbf{x})$. We have the following formulation: $\mathbf{x= x_{id} +R(x)}$, where $\mathbf{x}$ is the initial face feature, and $\mathbf{x_{id}}$ is the identity-dependent feature.

To reduce the mutual variations in the decomposed components, we propose a novel Decorrelated Adversarial Learning (DAL) algorithm that adversarially minimizes the correlation between $\mathbf{x_{id}}$ and $\mathbf{x_{age}}$. Specifically, a Canonical Mapping Module is introduced to find maximum correlations between $\mathbf{x_{id}}$ and $\mathbf{x_{age}}$, while the backbone and factorization module aims to reduce the correlation. In the meanwhile, $\mathbf{x_{id}}$ and $\mathbf{x_{age}}$ are learned by the identity and age classification signals respectively. Through the adversarial training, we wish the $\mathbf{x_{id}}$ and $\mathbf{x_{age}}$ will be sufficiently uncorrelated, and the age information in $\mathbf{x_{id}}$ can be significantly reduced.

Our major contributions are summarized as follows:

1. We propose a novel Decorrelated Adversarial Learning (DAL) algorithm based on the linear feature factorization, in order to regularize the learning of decomposed features. In this way, we wish to capture the ID-preserving while age invariant features for AIFR. To the best of our knowledge, this is the first work to introduce decorrelated adversarial feature learning to AIFR.

2. We present the Batch Canonical Correlation Analysis (BCCA), an extension of CCA in the fashion of stochastic gradient decent optimization. The proposed BCCA can be integrated to the deep neural networks for correlation regularization.

3. The proposed method has significantly improved the state-of-the-art performance on the AIFR datasets including MORPH Album2\cite{morph}, FG-NET\cite{fgnet} and CACD-VS\cite{cacd}, which strongly demonstrates its effectiveness.

%------------------------------------------------------------------------
\section{Related Work}
\textbf{Age-Invariant Face Feature Learning.}
Many prior studies\cite{mefa,lps_hfa,gsm,chen2013blessing,d1,d2,cacd,cacd2,hfa} in the literature extracted hand-craft features with heuristic methods. For example, the \cite{d1} developed a multi-feature discriminant analysis method with local feature descriptions. The \cite{hfa} proposed the hidden factor analysis (HFA) to model the feature factorization and reduce the age variations in identity-related features. The \cite{mefa} introduced an effective maximum entropy feature descriptor and a robust identity matching framework for AIFR. Several recent methods \cite{LFCNN,AECNN,oecnn} are mainly based on deep neural networks. The \cite{LFCNN} developed the Latent Factor guided Convolutional Neural Network (LF-CNN) to improve the HFA. The \cite{AECNN} introduced the Age Estimation guided CNN (AE-CNN) method for AIFR. The OE-CNN \cite{oecnn} proposed the orthogonal embedding decomposition such that the identity information is encoded in the angular space while the age information is represented in the radial direction. Our work presents a DAL algorithm with the linear residual decomposition.

\textbf{Canonical Correlation Analysis.}
Canonical Correlation Analysis (CCA) \cite{cca} is a well-known algorithm to measure the linear relationship between two multidimensional variables. Some previous works have introduced this method to face recognition in various scenarios. For example, the \cite{cca_2d3d} proposed a 2D-3D face matching method using the CCA. The \cite{cca_gong} developed a multi-feature CCA method for face-sketch recognition. Compared to these typical CCA based methods, our work presents the extension of CCA to deep neural network as a regularization method for AIFR.

\textbf{Adversarial Approaches.}
Generative adversarial networks (GAN) \cite{gan} have shown effective in various generative tasks, such as face aging \cite{caae,cgan,TNVP}, face super-resolution \cite{gan_sr,gan_sr2}, etc. Besides, the adversarial networks has also been explored to the improve the discriminative models. For example, the \cite{gan_det} utilized GAN to generate high-resolution of small faces in order to improve face detection. The \cite{uv_gan} developed an adversarial UV completion framework (UV-GAN) to solve the pose invariant face recognition problem. The \cite{gan_ae} proposed to learn the identity-distilled features and the identity-dispelled features in an adversarial autoencoder framework. The \cite{gan_htp} proposed an adversarial network to generate hard triplet feature examples. In this work, we propose a decorrelated adversarial learning method to significantly minimize the correlation between the decoupled components of identity and age, thus the identity-dependent features are age invariant.

%------------------------------------------------------------------------
\section{Method}
\subsection{Feature Factorization}

As faces contain intrinsic identity information and age information, they can be jointly represented by the identity-dependent features and the age-dependent features. Motivated by this, we design a linear factorization module that decomposes the initial features into these two unrelated components. Formally, given an initial feature vector $\mathbf{x} \in \mathbb{R}^{d}$ that extracted from an input image $\mathbf{p}$ by a backbone CNN $\mathcal{F}$ (i.e, $\mathbf x = \mathcal{F}(\mathbf{p})$), we define the linear factorization as follows:
\begin{equation}\label{1}
\begin{array}{c}
\mathbf{x} = \mathbf{x_{id}} + \mathbf{x_{age}},
\end{array}
\end{equation}
where $\mathbf{x_{id}}$ denotes the identity-dependent component, and $\mathbf{x_{age}}$ denotes the age-dependent component. We design a deep residual mapping module similar to \cite{dream} to implement this. Specifically, we obtain the age-dependent feature through a mapping function $\mathcal{R}$, and the residual part is regarded as the identity-dependent feature. We refer to this as Residual Factorization Module (RFM), which is formulated as:

\begin{equation}\label{2}
\begin{split}
\mathbf{x_{age}} &= \mathcal{R}~(\mathbf{x}), \\
\mathbf{x_{id}} &=  \mathbf{x} - \mathcal{R}~(\mathbf{x}). 
\end{split}
\end{equation}

\begin{figure}[t]
\begin{minipage}[b]{1.0\linewidth}
  \centering
  \centerline{\includegraphics[width=1.0\linewidth]{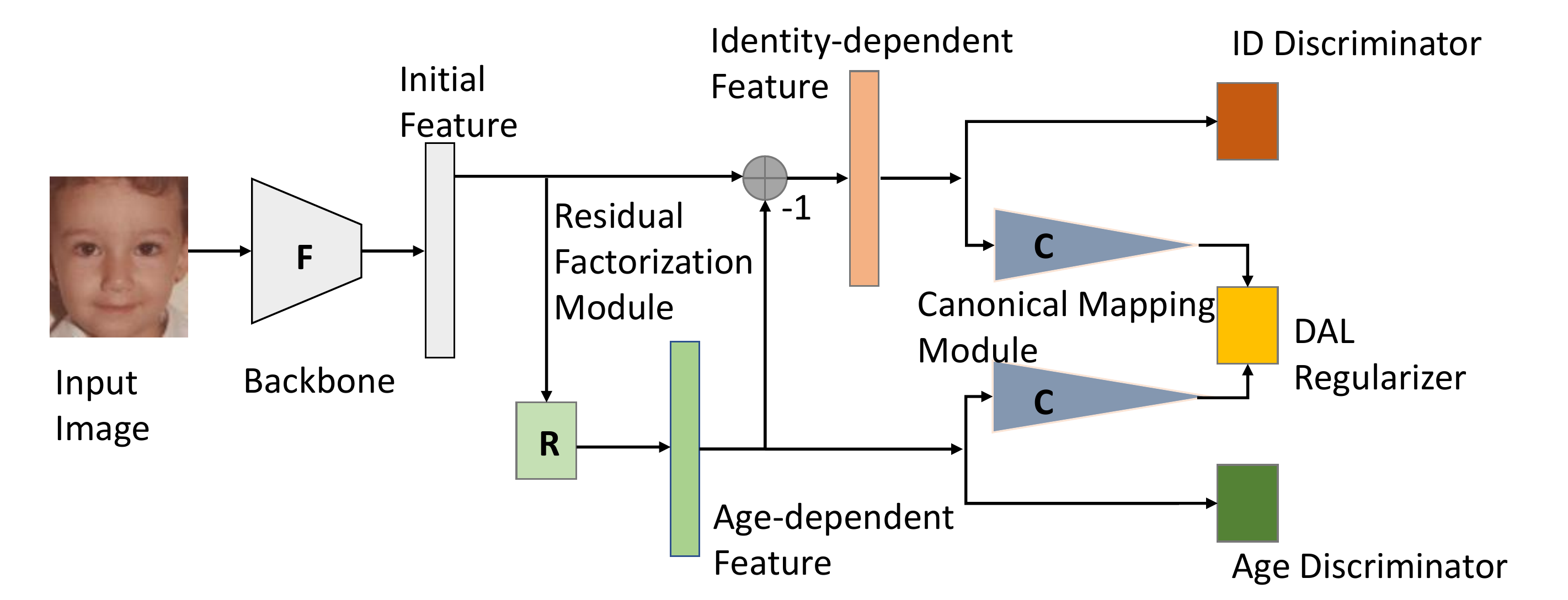}}
\end{minipage}
\caption{An overview of the proposed method. The initial features are extracted by a backbone net, followed by the residual factorization module. The two factorized components $x_{id}$ and $x_{age}$ are then used for classification and DAL regularization.}
\label{fig:framework}
\end{figure}

At testing stage, only the identity-dependent features are used for face recognition. It is desirable that $\mathbf{x_{id}}$ encodes the identity information while $\mathbf{x_{age}}$ draws the age variations. We simultaneously put the identity discriminating signal and the age discriminating signal onto these two decoupled features to respectively supervise the multi-task learning of these two components. 
Figure~\ref{fig:framework} shows the overall framework of our work. The resnet-like backbone extracts the initial features, upon which we build the residual module for feature factorization. Based on such factorization, we propose the Decorrelated Adversarial Learning, which is introduced in the following section.

%In particular, based on such factorization, we propose to reduce the correlation between the identity-specific components and age-specific components, which is introduced in the following section.

\subsection{Decorrelated Adversarial Learning}

Through feature factorization, it is crucial for AIFR that the $\mathbf{x_{id}}$ should be identity preserving and necessarily age-invariant. Unfortunately, the $\mathbf{x_{id}}$ and $\mathbf{x_{age}}$ practically have latent relationship with each other. For example, $\mathbf{x_{id}}$ and $\mathbf{x_{age}}$ may have high linear correlation with each other. Thus, the $\mathbf{x_{id}}$ may partially involve the age variation, which leads to negative effect on face recognition. On the other hand, the $\mathbf{x_{id}}$ and $\mathbf{x_{age}}$ should be mutually uncorrelated to force the non-trivial learning such that they both improve themselves.

To this end, we design a regularization algorithm that is helpful to reduce the correlation between the decomposed features, namely Decorrelated Adversarial Learning (DAL). The DAL basically calculates the canonical correlation between the paired features of the decomposed components.

Formally, given paired features $\mathbf{x_{id}, x_{age}}$, we design a linear Canonical Mapping Module (CMM) that maps 
%$x_{t}$ ($t \in \{id,age\}$) to the canonical variables $v_{t}$:
$\mathbf{x_{id}, x_{age}}$ to the canonical variables $\mathbf{v_{id}, v_{age}}$: 
\begin{equation}\label{cmm}
\forall t \in \{id,age\}: \mathbf{v_{t}}=\mathcal{C}(\mathbf{x_{t}}) = \mathbf{w_{t}^{T}}\mathbf{x_{t}}, 
\end{equation}
where the $\mathbf{w_{id}, w_{age}}$ are the learning parameters for canonical mapping. After that, we define the canonical correlation as:

\begin{equation}\label{ccor}
\begin{split}
%\rho &= \frac{(E[v_{id} - E[v_{id}]])(E[v_{age} - E[v_{age}])]]}{\sqrt[]{E[(v_{id} - E[v_{id}])^2]E[(v_{age} - E[v_{age}])^2]}} .\\
\rho &= \frac{\mathbf{Cov(\mathbf{v_{id}}, \mathbf{v_{age}})}}{\sqrt[]{\mathbf{Var(v_{id})Var(v_{id})}}}. \\
%(E[v_{id} - E[v_{id}]])(E[v_{age} - E[v_{age}])]]}{\sqrt[]{E[(v_{id} - E[v_{id}])^2]E[(v_{age} - E[v_{age}])^2]}} .\\
\end{split}
\end{equation}
Based on such definition, we first find maximum of $|\rho|$ by updating CMM with respect to $\mathbf{w_{id}, w_{age}}$, and then try to reduce the correlation by training the backbone and RFM. That is, on the one hand, we freeze $\mathcal{F},\mathcal{R}$ and train $\mathcal{C}$ in the canonical correlation maximizing process. On the other hand, we update $\mathcal{F},\mathcal{R}$ with $\mathcal{C}$ fixed in the feature correlation minimizing process. Obviously, they compete with each other playing a two-player min-max game during the adversarial training procedure. In this way, our goal is to minimize the correlation between $\mathbf{x_{id}, x_{age}}$ by always decreasing their maximum canonical correlation. In other words, the optimal feature projections having maximum correlation act as the primary target to be decorrelated. Thus, $\mathbf{x_{id}}$ and $\mathbf{x_{age}}$ learns continuously to have small correlation and finally they are significantly uncorrelated. 

Overall, the objective function for DAL is formulated as:
\begin{equation}\label{adv}
\begin{split}
\mathcal{L}_{DAL} = \min\limits_{\mathcal{F},\mathcal{R}} \max_{\mathcal{C}} (|\rho(\mathcal{C}(\mathbf{\mathcal{F}(\mathbf{p}}) - \mathcal{R}~(\mathbf{\mathcal{F}(\mathbf{p}})), \mathcal{C}(\mathcal{R}~(\mathbf{\mathcal{F}(\mathbf{p}})))|).
\end{split}
\end{equation}
We believe the strong decorrelation enhanced by DAL will encourage the $\mathbf{x_{id}}$ and $\mathbf{x_{age}}$ to be sufficiently invariant with each other. Importantly, this will improve robustness of $x_{id}$ for age-invariant face recognition.

\subsection{Batch Canonical Correlation Analysis}

In contrast to the typical canonical correlation analysis (CCA) methods, our work introduces the canonical correlation Analysis (BCCA) based on stochastic gradient decent (SGD) optimization. Since the correlation statistics on the entire dataset is practically impossible, we follow similar strategy of batch normalization \cite{bn} to compute the correlation statistics based on mini-batches. Thus, it naturally suits the deep learning framework.

Given a mini-batch size of $m$, we have two sets of the decomposed features: $B_{id} = \{x_{id}^{1, ..., m}\}$  and $B_{age} = \{x_{age}^{1, ..., m}\}$. Thus, the canonical correlation can be written as:

\begin{equation}\label{bcc}
\mathbf{\rho = \frac{\frac{1}{m}\Sigma_{i=1}^{m}{(v_{id}^{i} - \mu_{id})(v_{age}^{i} - \mu_{age})}}{\sqrt[]{\sigma_{id}^2+\epsilon}~~\sqrt[]{\sigma_{age}^2+\epsilon}}}.\\
\end{equation}
Here, the $\mathbf{\mu_{id}}$ and $\sigma_{id}^2$ are the mean and variance of $\mathbf{v_{id}}$ respectively, and similarly for $\mu_{age}$ and $\sigma_{age}^2$. The $\epsilon$ is a constant parameter for numerical stability.

Equation \ref{bcc} serves as the objective function for BCCA and we leverage the SGD based algorithm to optimize it. Note that the canonical correlation $|\rho|$ is demanded to be necessarily maximized when updating the $\mathcal{C}$., while being minimized when training the $\mathcal{F},\mathcal{R}$. The derivation of gradients are:

\begin{equation}\label{grad}
\begin{split}
\frac{\mathbf{\partial{\rho}}}{\mathbf{\partial{{v_{id}^{i}}}}} &= \frac{1}{m}\mathbf{(\frac{v_{age}^{i} - \mu_{age}}{\sqrt[]{\sigma_{id}^2+\epsilon}~~\sqrt[]{\sigma_{age}^2+\epsilon}} - \frac{(v_{id}^{i} - \mu_{id})\cdot\rho}{\sigma_{id}^2+\epsilon})}, \\
\frac{\partial{\rho}}{\mathbf{\partial{{v_{age}^{i}}}}} &= \frac{1}{m}\mathbf{(\frac{v_{id}^{i} - \mu_{id}}{\sqrt[]{\sigma_{id}^2+\epsilon}~~\sqrt[]{\sigma_{age}^2+\epsilon}} - \frac{(v_{age}^{i} - \mu_{age})\cdot\rho}{\sigma_{age}^2+\epsilon})}. \\
\end{split}
\end{equation}

Thus, the optimization consists of a forward propagation that outputs the $\rho$, and a backward propagation that calculate the gradients for updating. The detailed learning algorithm of BCCA is described in Algorithm \ref{alg:bcca}.

\begin{algorithm}[htb]
  \caption{  Learning algorithm of BCCA for each iteration.}
  \label{alg:bcca}
  \begin{algorithmic}[1]
    \Require
     $B_{id} = \{x_{id}^{1, ..., m}\}$; $B_{age} = \{x_{age}^{1, ..., m}\}$;
    %\Parameter: $B_{id}$
    \Ensure
      the canonical correlation $\rho$ for forward pass;
      the gradients for backward pass.
 
    \For {each $t \in \{id,age\}$}   
    \State CMM forward: $v_{t}^{i} = w_{t}^{T}x_{t}^{i}$ for $i=1 \dots m$; %,   t \in {id,age}$ ;
    \State Compute means: $\mu_{t} = \frac{1}{m}\Sigma_{i=1}^{m}{v_{t}^{i}}$; 
    \State Compute variances: $\sigma_{t}^2 = \frac{1}{m}\Sigma_{i=1}^{m}{(v_{t}^{i} - \mu_{t})^2}$; 
    \EndFor
    \State Forward propagation: Compute $\rho$ with Equation \ref{bcc}. 
    \For {each $t \in \{id,age\}$} 
	\State Compute $\frac{\partial{\rho}}{\partial{{v_{t}}}}$ with Equation \ref{grad};
%	\State Compute $\frac{\partial{\rho}}{\partial{{v_{age}}}}$ with Equation \ref{grad}; 
    \State CMM backward: $\frac{\partial{L}}{\partial{{x_{t}^{i}}}} = w_{t}^{i}\frac{\partial{L}}{\partial{{v_{t}^{i}}}}$; for $i=1 \dots m$;
	\State CMM backward: $\frac{\partial{L}}{\partial{{w_{t}^{i}}}} = x_{t}^{i}\frac{\partial{L}}{\partial{{v_{t}^{i}}}}$; for $i=1 \dots m$;
	\EndFor      
  \end{algorithmic}
\end{algorithm}

\subsection{Multi-task Training}
In this section, we describe the multi-task training strategy to supervise the learning of the decomposed features. As shown in Figure \ref{fig:framework}, there are three basic supervision modules: age discriminator, identity discriminator and DAL regularizer.

\textbf{Age Discriminator.} 
For the learning of age information, we feed $\mathbf{x_{age}}$ into an age discriminator to ensure the age discriminating information. Since age labels are rough with uncertain noises in practice, we follow \cite{hfa,LFCNN} and perform classifications on ages by dividing them into different groups. We use the softmax layer with cross-entropy loss for the age classification.

\textbf{Identity Discriminator.}
Following the recent \cite{cosface,amsm}, we utilize the CosFace loss to supervise the learning of $\mathbf{x_{id}}$ and ensure the  identity-preserving information. The CosFace loss is formulated as:

\begin{equation}\label{idloss}
 \mathcal{L}_{ID} = \frac{1}{N}\sum_{i}{-\log{\frac{e^{s (\mathbf{\cos(\theta_{{y_i}, i})} - m)}}{e^{s (\mathbf{\cos(\theta_{{y_i}, i})} - m)} + \sum_{j \neq y_i}{e^{s \mathbf{\cos(\theta_{j, i})}}}}}},
\end{equation}
where $N$ is the number of identities, $y_i$ is the corresponding identity label, $\mathbf{cos(\theta_j,i)} = \mathbf{{\frac{W_j^{T}}{\Vert{W_j}\Vert}}\cdot {\frac{x_{id}^{i}}{\Vert{x_{id}^{i}}\Vert}}}$ is the cosine of angle between the i-th feature $\mathbf{x_{id}^{i}}$ and the j-th weight vector $\mathbf{W_j}$ of the classifier. The $m$ a constant margin term controlling the cosine margin and the $s$ is a constant scaling factor $s$. The CosFace loss aims to introduce much strict constraints to the identity classification such that the learned features are encouraged to be separated by a margin between different identities. A properly large $m$ will encourage powerful discriminating information in the learned features for face recognition.

\textbf{DAL Regularizer.}
The proposed DAL regularization also participants the joint supervision to guide the feature learning such that the correlations between the paired decomposed features can be significantly reduced. Through the joint supervision, the model simultaneously learns to encourage both the discriminating power of $x_{id}$, $x_{age}$, and decorrelation information between of these two decomposed components.

In summary, the training is supervised by the following combined multi-task loss:

\begin{equation}\label{mulloss}
\mathcal{L} = \mathcal{L}_{ID}(\mathbf{x_{id}}) + \lambda_1 \mathcal{L}_{SM}(\mathbf{x_{age}}) + \lambda_2 {\mathcal{L}_{DAL}(\mathbf{x_{id}},  \mathbf{x_{age}})},
\end{equation}
where $L_{ID}$ denotes the CosFace loss, $L_{SM}$ denotes the softmax with cross-entropy loss, $\lambda_1$ and $\lambda_2$ are scalar hyper-parameters to balance these three losses.
In the testing phase, we extract the identity-dependent features $x_{id}$ for AIFR evaluations.

\subsection{Discussion}

The proposed method has the following advantages. First, the DAL regularization on features is helpful to encourage the uncorrelated and co-invariant information between the decomposed components. Related works such as HFA\cite{hfa}, LF-CNN\cite{LFCNN} and OE-CNN\cite{oecnn} have neglected the underlying correlation. Instead, we aim to minimize the classification error as well as the correlation effect simultaneously. Second, the BCCA provides an extension of CCA that is inserted to the deep learning framework such that the entire model can be trained in an end-to-end process. Finally, our method can be easily generalized to other components factorization model, such as pose, illumination, emotion, etc. To the best of our knowledge, we are the first to develop the decorrelated adversarial regularization framework to AIFR.

%------------------------------------------------------------------------
\section{Experiments}

\subsection{Implementation Details}
\textbf{Network Architecture.} (1) Backbone: our backbone network is a 64-layer CNN similar to \cite{oecnn} . It consists of 4 stages with respectively 3, 4, 10, 3 stacked residual blocks. Every residual block has 3 stacked units of ``3x3 Conv + BN + ReLU". Finally, a FC layer outputs the initial face features of 512 dimension. (2)Residual factorization Module (RFM): the initial face features are mapped to form the age-dependent feature through 2 ``FC +ReLU'', and the residual part is regarded as the identity-dependent feature. (3) Age discriminator: we stack 3 ``FC +ReLU'' upon $\mathbf{x_{age}}$, and perform age classification. (4) Identity discriminator: we directly use $\mathbf{x_{id}}$ for identification by CosFace loss. (5) DAL regularizer: we feed the $\mathbf{x_{age}}$ and $\mathbf{{x_{id}}}$ into the FC layers respectively and output their linear combinations, which are then used for the  BCCA calculation and optimization.

\begin{figure}[t]
\begin{minipage}[b]{1\linewidth}
  \centering
  \centerline{\includegraphics[width=1.0\linewidth]{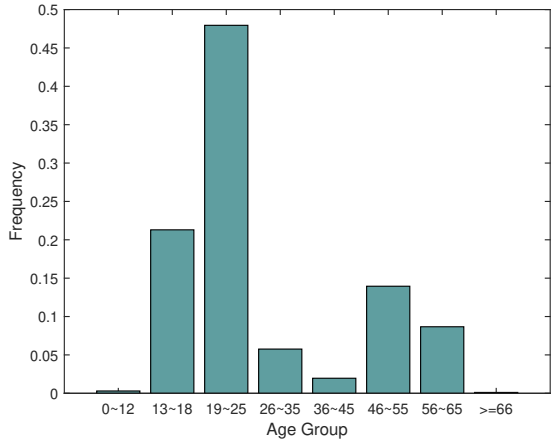}}
\end{minipage}
\caption{ The age distribution of our small training dataset. It contains 0.5M face images covering large age variations. 
%We use this dataset to train our models for visualization analysis and benchmark comparisons. 
}
\label{fig:age_hist}
\end{figure}

\textbf{Data Preprocessing.}
We use MTCNN \cite{mtcnn} to detect face areas and facial landmarks on both the training and testing sets. Then, similarity transformation is performed according to the 5 facial key points (two eyes, nose and two mouth corners) in order to crop the face patch to 112$\times$96 . Finally, each pixel ([0,255]) of the cropped face patch is normalized by subtracting 127.5 then divided by 128.

\begin{figure*}[t]
\begin{minipage}[b]{1\linewidth}
  \centering
  \centerline{\includegraphics[width=1.0\linewidth]{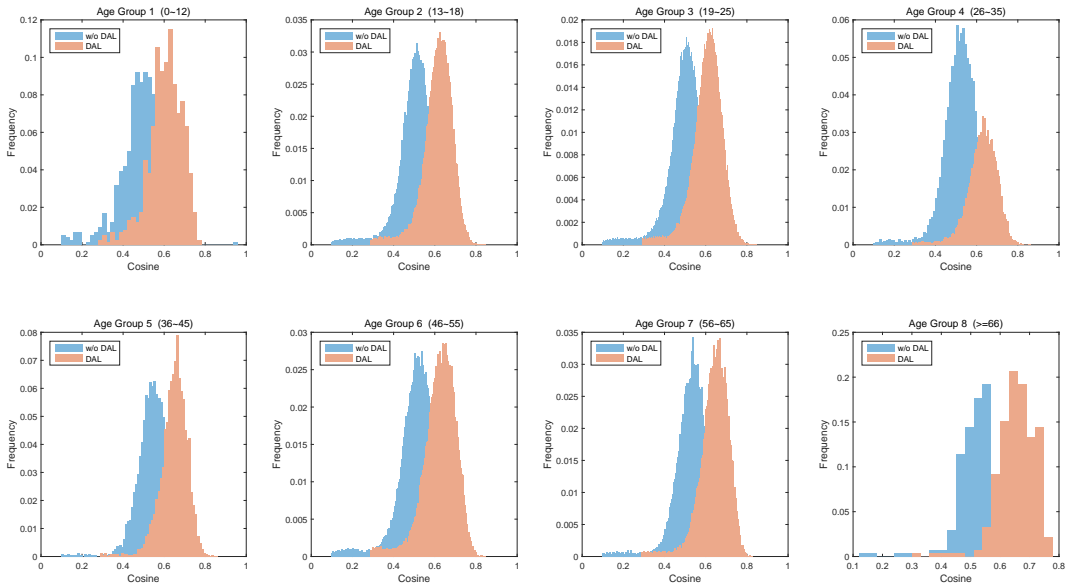}}
\end{minipage}
\caption{ The distribution of the cosine similarity between features and their class centers at different age groups. Our DAL model consistently increases the cosine similarity compared against the baseline model without DAL across all the age groups, which demonstrates the effectiveness of our model to encourage less  intra-identity variations. Best viewed in colors.}
\label{fig:hist}
\end{figure*}

\textbf{Training Details.}
Our training data includes the Cross-Age Face (CAF) dataset provided by \cite{oecnn} and other common face datasets such as CASIA-WebFace \cite{webface}, VGG Face \cite{vggface} and celebrity+ \cite{celeb}. It totally contains about 1.7M images from 19.9k individuals, which is similar to \cite{oecnn}. Meanwhile, we build a subset containing about 0.5M images from 12k individuals following \cite{oecnn} in order to conduct fair experimental comparisons. We refer to this subset as \textit{small training dataset} and our whole training dataset as \textit{large training dataset} for clarify.
We adopt the pre-trained age estimation model \cite{imdb2} to generate predicted age labels for the face images of the entire training set. Note that only those predicted ages with relatively high confidence (i.e. more likely to be true label) are considered valid and will participate the age-classification. After that, the predicted ages are divided into 8 groups: 0-12, 13-18, 19-25, 26-35, 36-45, 46-55, 56-65, $\geq 66$. The grouped age labels are then used for the age-classification training.
The joint supervision in Equation \ref{mulloss} guides the DAL　training process in an adversarial manner. More specifically, in an adversarial loop, we alternately run the canonical correlation maximizing process for 20 iterations and then change to feature correlation minimizing process for 50 iterations. The empirically setting of hyper-parameters $\lambda_1$ and $\lambda_2$ in Equation \ref{mulloss} are: $\lambda_1 = 0.1$, $\lambda_2 = 0.1$, $m = 0.35$, $s = 64$.
All our experiment models are trained through stochastic gradient descent (SGD), with batch size of 512. The whole training procedure is about 40-th epochs and the learning rate is initially set to 0.1 and reduced by a factor of $0.1$ at 22-th, 33-th, 38-th epoch.

\textbf{Testing Details.}
We conduct evaluation experiments on the well-known public AIFR face datasets: FG-NET\cite{fgnet}, MORPH Album 2\cite{morph} and CACD-VS\cite{cacd}. In the testing process, we extract the identity-dependent features and concatenate features of the original image and the flipped image to form the final representation. The cosine similarity of these representations are then used to conduct face verification and identification.

%% ~~~~~~~~~~~~ Ablation Study

\begin{table}
\footnotesize
\begin{center}
\begin{tabular}{c|c|c|c|c|c}
\hline
Model    	& \tabincell{c}{FG-NET \\(MF1)}    & \tabincell{c}{FG-NET \\(MF2)}  & \tabincell{c}{FG-NET \\(leave-\\one-out)} & \tabincell{c}{MORPH\\Album 2} & \tabincell{c}{CACD\\-VS}\\ 
\hline\hline
Baseline       & 55.86\%      &    		58.85\%			& 93.4\%	& 98.21\%   & 99.07\%	\\
+Age       & 55.84\%      &   58.64\% 		& 93.6\%	& 98.11\%   & 99.05\%	\\
\hline
\textbf{+Age+DAL}           & \textbf{57.92\%}        &  \textbf{60.01\%} 		& \textbf{94.5\%}	
& \textbf{98.93\%}      &   \textbf{99.40\%}	\\
\hline

\end{tabular}
\end{center}
\caption{ Comparison of our method against the baseline models. The evaluation results are rank-1 face identification rates on FG-NET, under protocols of MF1, MF2, and leave-one-out.}
\label{ablation}
\end{table}

\subsection{Ablation Study}

In this subsection, we study the different variants of the proposed models to show the effectiveness of our method. 

\textbf{Visualization of Cosine Similarity.} 
For a better understanding of the DAL and its ability to improve the identity-preserving information,  we conduct an experiment to visualize the cosine similarities across different age groups. Given the learned identity-dependent features $\mathbf{x_{id}}$, we first calculate their class centers by clustering every identity in the identity feature space, and then compute the cosine similarity between each sample and its class center. After that, we plot the distribution of cosine similarity across different age groups. In this study, we conduct such visualization analysis on the small training dataset which contains 0.5M face images covering various age differences. Figure \ref{fig:age_hist} shows the age distribution of this dataset. We present a comparison between the ``w/o DAL'' model (trained by the joint supervision signals of age and identity but without DAL) and our proposed DAL model. As shown in Figure \ref{fig:hist}, compared against the ``w/o DAL'' model, the DAL model consistently increases the cosine similarity between $\mathbf{x_{id}}$ and its class center across all the age groups. This observation proves that our method encourages features to have small intra-identity variations and thus the samples of the same identity but different ages are pulled together. Thus, the discriminating power of the learned identity features can be effectively improved by the proposed DAL method.

%Besides, one can observe that the cosine similarity increases of child ages or elder ages are relatively higher than those of other age phases, which further demonstrates the effectiveness of DAL in learning.

\textbf{Quantitative Evaluation.} To show the impact of the joint learning framework with our proposed DAL method, we conduct the ablative evaluations on several public AIFR datasets including FG-NET, MORPH Album 2 and CACD-VS. Moreover, we also test our models on FG-NET following the protocols of Megaface challenge 1 (MF1) \cite{mf1} and Megaface challenge 2 (MF2) \cite{mf2}. Both the MF1 and the MF2 include an additional distractor set respectively that contains 1 million face distractors, making the benchmarks much more difficult. The MF2 provides a training dataset such that all the evaluation methods should be trained on the same dataset and without any additional training data.  We consider the following models for ablative comparison in this study: (1) Baseline: the baseline model is trained by the identification loss only and without any extra age supervision. (2) +Age: this model is trained by the joint supervision of the identification signal and the age classification signal. (3) +Age+DAL: our proposed model that is trained simultaneously by the DAL regularization and the joint supervision signals. As reported in Table \ref{ablation}, without DAL the joint supervision model achieves comparable results with the baseline model. On the contrary, our ``+Age+DAL'' model improves the performance of FG-NET on all the schemes. The improvement on FG-NET with the scheme of MF2 is relatively limited compared with that of MF1 and 'leave-one-out', mainly due to the less aging variations of MF2 training dataset. Nevertheless, the consistently performance improvement demonstrates the effectiveness of our method. Moreover, our method improves the baseline models by more than 0.7\% on  MORPH Album 2, and more than 0.3\%  on CACD-VS, which are remarkable improvements at the high accuracy level above 98\% and 99\%.

%% `~~~~~~~~~ Morph

\begin{table}[t]
\small
\begin{center}
\begin{tabular}{c|c|c}
\hline
Method	& \#Test Subjects			& Rank-1\\
\hline\hline
HFA \cite{hfa} 		& 10,000			& 91.14\% \\
CARC \cite{cacd}	& 10,000				& 92.80\% \\
MEFA \cite{mefa}	& 10,000				& 93.80\% \\
MEFA+SIFT+MLBP \cite{mefa}	& 10,000	& 94.59\% \\
LPS+HFA \cite{lps_hfa}	& 10,000		& 94.87\% \\
LF-CNNs \cite{LFCNN}	& 10,000			& 97.51\% \\
OE-CNNs		&	 10,000						& 98.55\% \\
\hline
\textbf{Ours}		&	 10,000			& \textbf{98.93\%} \\
\hline
GSM \cite{gsm}		& 3,000			& 94.40\% \\
AE-CNNs \cite{AECNN}	& 3,000			& 98.13\% \\
OE-CNNs \cite{oecnn}				& 3,000			& 98.67\% \\
\hline
\textbf{Ours} 				& 3,000			& \textbf{98.97\%} \\

\hline
\end{tabular}
\end{center}
\caption{Evaluation results on the MORPH Album 2 dataset.}
\label{tabel:MORPH-EXP}
\end{table}

%% `~~~~~~~~~ CACD

\begin{table}[t]
\begin{center}
\begin{tabular}{c|c|c}
\hline
Method 													& Acc.					& AUC.\\
\hline
\hline
High-Dimensional LBP \cite{chen2013blessing}		& 81.6\%				& 88.8\% \\
HFA \cite{hfa} 									& 84.4\%				& 91.7\% \\
CARC \cite{cacd}									& 87.6\%				& 94.2\% \\
LF-CNNs \cite{LFCNN}								& 98.5\%				& 99.3\% \\
Human, Average \cite{cacd2}						& 85.7\%				& 94.6\% \\
Human, Voting \cite{cacd2} 						& 94.2\%				& 99.0\% \\
%\hline\hline
Softmax										& 98.4\% 				& 99.4\% \\
A-Softmax										& 98.7\% 				& 99.5\% \\
OE-CNNs	\cite{oecnn}										& 99.2\% 		& 99.5\% \\
\hline
\textbf{Ours}									& \textbf{99.4\%} 		& \textbf{99.6\%} \\

\hline
\end{tabular}
\end{center}
\caption{Evaluation results on the CACD-VS dataset.}
\label{tabel:CACD-EXP}
\end{table}

\subsection{Experiments on the MORPH Album 2 Dataset}

The MORPH Album 2 dataset consists of 78,000 face images of 20,000 individuals across different ages. For fair comparison, we follows \cite{oecnn} and conduct evaluations under two benchmark schemes where the testing set consists of 10,000 subjects and 3,000 subjects respectively. In the testing sets, two face images of each subjects with the largest age gaps are selected to compose the probe set and the gallery set. We train the model with our proposed DAL on the large training dataset(1.7M images). Note that we have not conducted any training or finetuning on the MORPH Album 2. 

In this experiment, we compare our DAL model against the recently AIFR algorithms in the literature. As shown in Table \ref{tabel:MORPH-EXP}, the proposed method has effectively improved the rank-1 identification performance. Particularly, our method outperforms the recent top-performing AIFR methods by a clear margin, setting new state-of-the-art on the MORPH Album 2 database.

\subsection{Experiments on the CACD-VS Dataset}

As a public released dataset for AIFR, the CACD dataset is composed of 163,446 images from 2,000 celebrities with age variations. The collected face images also include different illumination, various poses and makeup. The subset CACD-VS consists of 4000 face image pairs for face verification, and the face pairs are divided into 2,000 positive pairs and 2,000 negative pairs. In our experiment, we strictly follow \cite{cacd,oecnn} to perform the 10-fold cross-validation for fair comparisons. 
We use the same trained models in Sec 4.3 to evaluate the performance on the CACD-VS Dataset. Table \ref{tabel:CACD-EXP} shows the verification accuracy of our models compared against the other state-of-the-art AIFR methods. Not surprisingly, the proposed DAL model obtains consistent improvement over the prior methods, demonstrating the superiority of our method again.

%% ~~~~~leave one out

\begin{table}[t]
%\small
\begin{center}
\begin{tabular}{c|c}
\hline
Method 								& Rank-1\\
\hline\hline
Park et al. \cite{g3} (2010)					& 37.4\% \\
Li et al. \cite{d1} (2011)					& 47.5\% \\
HFA \cite{hfa} (2013)				& 69.0\% \\
MEFA \cite{mefa} (2015)		& 76.2\% \\
CAN \cite{can}						& 86.5\% \\
LFCNNs \cite{TNVP}					& 88.1\% \\
\hline
\textbf{Ours} 						& \textbf{94.5\%} \\
\hline
\end{tabular}
\end{center}
%\caption{Performance of different published approaches on FG-NET. Note that the experiments are conducted under the protocol of MF1 \cite{mf1}.}
\caption{Evaluation results on the FG-NET dataset under the protocol of leave-one-out.}
\label{tabel:FGNET-loo}
\end{table}

%% `~~~~~~~~~ MF1

\begin{table}[t]
%\small
\begin{center}
\begin{tabular}{c|c|c}
\hline
Method 			& Protocol 			& Rank-1\\
\hline\hline
FUDAN-CS\underline{ }SDS \cite{fudan}	& Small	& 25.56\% \\
SphereFace \cite{sphereface}		& Small							& 47.55\% \\
TNVP \cite{TNVP}			& Small						& 47.72\% \\
%SIAT\underline{ }MMLAB		& Small						& 55.30\% \\
%FaceNet v8 \cite{facenet} & Large						& 74.59\% \\
%\hline\hline
%Softmax					& Large						& 44.09\% \\
%A-Softmax				& Large							& 50.69\% \\
%\hline\hline
Softmax					& Small						& 35.11\% \\
A-Softmax				& Small						& 46.77\% \\
%单patch
OE-CNNs \cite{oecnn} & Small						& 52.67\% \\
%多patch
%\OE-CNNs (3-patch ensemble)			& Small						& \textbf{58.21\%} \\
% \textbf{OE-CNNs}			& Large						& \textbf{59.08\%} \\
\hline
%CosFace 					&small						& 56.06\% \\
\textbf{Ours} 					&small						& \textbf{57.92\%} \\
\hline
\end{tabular}
\end{center}
%\caption{Performance of different published approaches on FG-NET. Note that the experiments are conducted under the protocol of MF1 \cite{mf1}.}
\caption{Evaluation results on the FG-NET dataset under the protocol of MF1.}
\label{tabel:FGNET-EXP}
\end{table}

%~~~~~~~~~~~~~~~~ MF2

\begin{table}[t]
\small
\begin{center}
\begin{tabular}{c|c|c}
\hline
Method 			& Protocol 						&Rank-1\\
\hline\hline
GRCCV	& Large	& 21.04\% \\
NEC	& Large	& 29.29\% \\
3DiVi	& Large	& 35.79\% \\
GT-CMU-SYSU	& Large	& 38.21\% \\
%\hline\hline
%Softmax					& Large						& 44.09\% \\
%A-Softmax				& Large							& 50.69\% \\
OE-CNNs \cite{oecnn}			& Large						& 53.26\% \\
\hline
%CosFace (single-patch)			& Large						& 59.14\% \\
\textbf{Ours}			& Large				& \textbf{60.01\%} \\

\hline
\end{tabular}
\end{center}
\caption{Evaluation results on the FG-NET dataset under the protocol of MF2.}
\label{tabel:FGNET-EXP2}
\end{table}

\subsection{Experiments on the FG-NET Dataset}

The FG-NET comprises 1002 face images from 82 individuals, with ages from 0 to 69. The dataset includes lots of face images at the age phase of the child and the elderly. We conducted experiments under three evaluation schemes for benchmark comparison: leave-one-out, MegaFace challenge 1 (MF1) and MegaFace challenge 2 (MF2).

\textbf{Evaluation with leave-one-out.}
We directly use the DAL model trained on the small training set (0.5M images) and test on the FG-NET dataset. The evaluation is conducted by leave-one-out. It is noticeable that we have not used any data of FG-NET for training or finetuning. The performance comparisons are given in Table \label{tabel:FGNET-loo}. We can see that our method has improved the priors \cite{hfa} by a significant margin.

\textbf{Evaluation with MF1.}
The MF1 \cite{mf1} contains 1 million distractor images from 690K different individuals. According to \cite{mf1}, evaluations are conducted under the two protocols: large or small training set. The training set less than 0.5M is considered small. We strictly follow the protocol of small training set to train the model and conduct evaluations on FG-NET. The experimental results are reported in Table \ref{tabel:FGNET-EXP}. The performance improvement over the other methods strongly demonstrates the effectiveness of the proposed DAL method. %Our results of 3-patch ensembles has further improved the best AIFR performance in the literature.

\textbf{Evaluation with MF2.} 
We also conducte experiments on the MF2 \cite{mf2}, which has 1 million distractors as well. But
the distractors of MF1 and MF2 are totally different. Unlike the MF1, the MF2 requires that all the models should be trained on the same training set, thus yields very fair comparisons. The training set provided by MF2 contains 4.7 million faces from 672K identities. Following this protocol, we train our models and conduct evaluations on the MF2. Table \ref{tabel:FGNET-EXP2} shows the performance comparisons between ours and the previous methods. Again, our DAL method significantly improves the identification accuracy and set new state-of-the-art on the MF2 dataset.

\subsection{Experiments on the General Face Recognition Datasets}

\begin{table}[t]
\small
\begin{center}
{
\small
\begin{tabular}{c | c |c}
\hline
Method & LFW & 
\tabincell{c}{MF1-Facescrub}\\
\hline\hline
SphereFace\cite{sphereface} & 99.42\% & 72.73\% \\
\hline
CosFace\cite{cosface} & 99.33\% & 77.11\% \\
\hline
OE-CNNs\cite{oecnn} & 99.35\% & N/A \\
\hline
\textbf{Ours} & \textbf{99.47\%} & \textbf{77.58\%} \\
\hline
\end{tabular}
}
\end{center}
\caption{Evaluation results on LFW and MF1-Facescrub datasets. The reported results are verification rates for LFW, and rank-1 identification rates for MF1-Facescrub.}
\label{tabel:GFR-EXP}
\end{table}

To compare against the state-of-the-art methods in General Face Recognition(GFR), we further conduct experimental evaluations on the LFW and the MegaFace Challenge 1 Facescrub (MF1-Facescrub) datasets. The LFW \cite{lfw} is a public benchmark for GFR that has 13,233 face images from 5,749 subjects. The MF1-Facescrub \cite{mf1} includes the Facescrub (containing 106,863 face images from 530 celebrities) as a probe set and contains a million distractors in the gallery set. Following the evaluation procedure in OE-CNNs \cite{oecnn}, our training data contains 0.5M images that are the same as OE-CNNs \cite{oecnn}. Table \ref{tabel:GFR-EXP} reports the verification rate on LFW and the rank-1 identification rate in MF1-Facescrub. Our model outperforms the \cite{oecnn} as well as the state-of-the-art General Face Recognition (GFR) models \cite{sphereface,cosface} on both datasets, which demonstrates the strong generalization ability of our proposed approach.

%------------------------------------------------------------------------
\section{Conclusion}

In this paper, we proposed the decorrelated adversarial learning (DAL) method for AIFR. Our model learns to minimize the correlation between the paired decomposed features of identity and age in an adversarial process. We presented the Batch Canonical Correlation Analysis (BCCA) algorithm as an extension of CCA in deep learning. Besides DAL, we simultaneously trained the model with the joint supervision of identification and age classification. In the testing, only the identity features were used for face recognition. The evaluations conducted on the AIFR benchmarks demonstrate the superiority of our proposed method.

{\small
\bibliographystyle{ieee}
\bibliography{egbib}
}

\end{document}